\pdfoutput=1

\documentclass[11pt]{article}

\usepackage[]{EMNLP2023}
\usepackage{times}
\usepackage{latexsym}

\usepackage[T1]{fontenc}

\usepackage[utf8]{inputenc}

\usepackage{microtype}

\usepackage{inconsolata}

\usepackage{adjustbox}
\usepackage{booktabs}

\title{Humanoid Agents: Platform for Simulating Human-like Generative Agents}

\author{
  Zhilin Wang* \\
  University of Washington \\
  and NVIDIA \\
  \texttt{zhilinw@uw.edu} \\\And
  Yu Ying Chiu* \\
  University of Washington \\
  \texttt{kellycyy@uw.edu} \\\And
  Yu Cheung Chiu \\
  The University of Hong Kong \\
  \texttt{jackycyc@connect.hku.hk} \\
}

\begin{document}
\maketitle
\begin{abstract}

Just as computational simulations of atoms, molecules and cells have shaped the way we study the sciences, true-to-life simulations of human-like agents can be valuable tools for studying human behavior. We propose Humanoid Agents, a system that guides Generative Agents to behave more like humans by introducing three elements of System 1 processing: Basic needs (e.g. hunger, health and energy), Emotion and Closeness in Relationships. Humanoid Agents are able to use these dynamic elements to adapt their daily activities and conversations with other agents, as supported with empirical experiments. Our system is designed to be extensible to various settings, three of which we demonstrate, as well as to other elements influencing human behavior (e.g. empathy, moral values and cultural background). Our platform also includes a Unity WebGL game interface for visualization and an interactive analytics dashboard to show agent statuses over time. Our platform is available on \url{https://www.humanoidagents.com/} and code is on \url{https://github.com/HumanoidAgents/HumanoidAgents}.


\end{abstract}

\section{Introduction}

\def\thefootnote{*}\footnotetext{Equal Contribution}\def\thefootnote{\arabic{footnote}}


Ushered in by the landmark paper on Generative Agents  \citep{park2023generative}, the promise of modelling conceivable human behavior using advanced NLP systems has sparked the imagination of many. Generative Agents plan activities over a day, execute them at each time-step and adapt their plans based on observations of their environment. While this approach can generate seemingly believe-able activities to external observers, this process does not fully resemble how humans think. Most of us do not create plans well in advance, and then meticulously and precisely carry out those plans in day-to-day life. Instead, we constantly adapt our plans to how we feel on the inside, in addition to changes in our physical environment.

\begin{figure}
\includegraphics[width=\columnwidth]{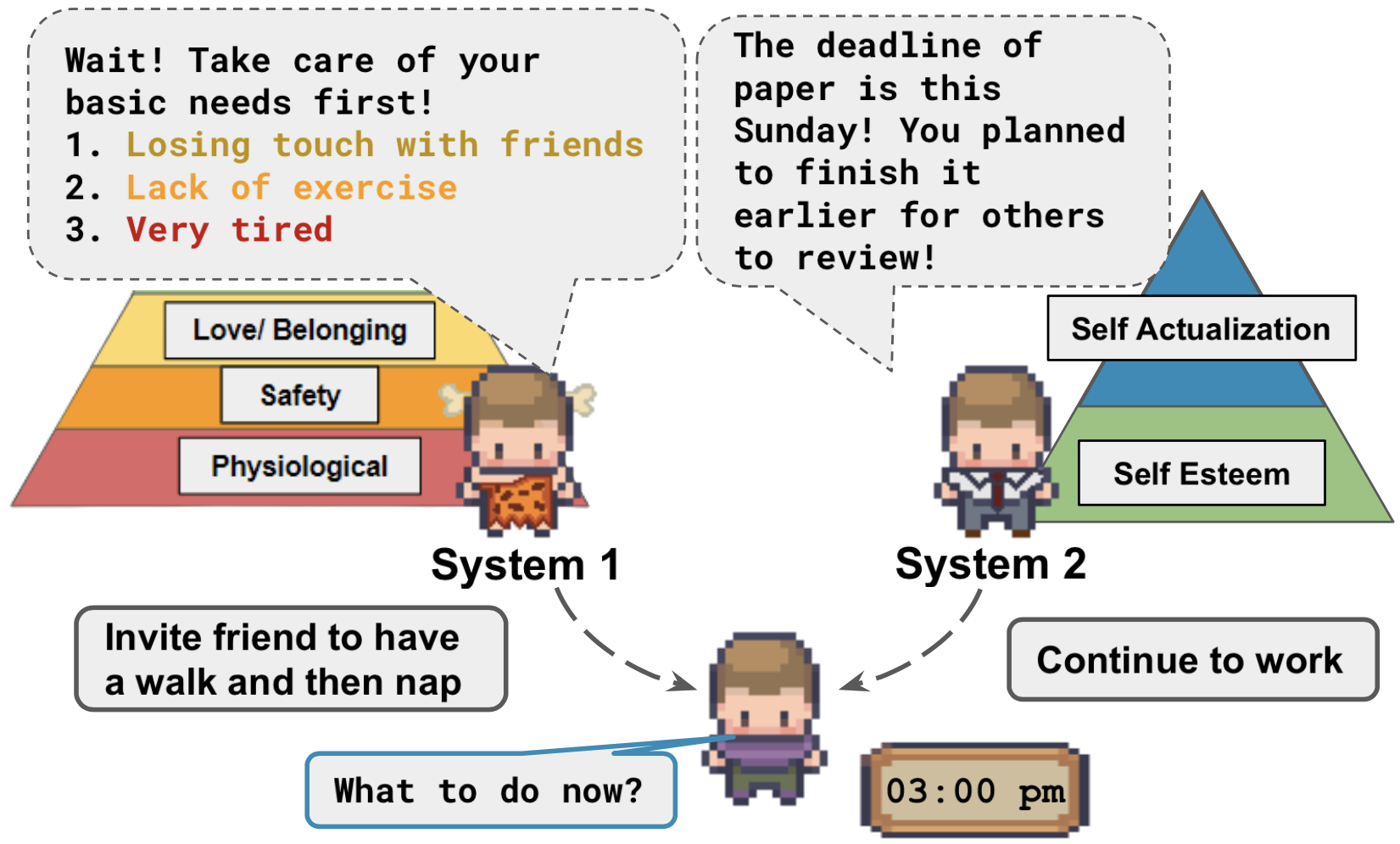}
\caption{Humanoid Agents are guided by both System 1 thinking to respond to their embodied conditions such as their basic needs and System 2 thinking involving explicit planning.}
\label{fig:front_page}
\end{figure}
To mitigate this shortcoming, we draw inspiration from psychology to propose Humanoid Agents. \citet{kahneman2011thinking} suggests that humans have two complementary processes for thinking: System 1 is intuitive, effortless and instantaneous while System 2 is logical, intentional and slow. Generative Agents focus on System 2 thinking at the cost of System 1. To better guide the behavior of Humanoid Agents using System 1, we introduce three aspects of System 1 that can influence their behavior: Basic needs, Emotion and the Closeness of their social relationship with other agents.

Basic needs refer to intrinsic needs that humans have for survival \citep{maslow:motivation}. To appropriately model human behavior, agents need to  interact with others, maintain their health and rest. Failing to do so adequately, agents will receive negative feedback comprising loneliness, sickness and tiredness, as illustrated in Fig. \ref{fig:front_page}. System 2 planning alone can implicitly account for activities to meet these needs (e.g. planning time for rest) but without feedback from System 1, agents cannot adapt by having a nap at 3pm if tired but bedtime is planned for midnight. Similarly, a realistic model of human behavior needs to consider the agents' emotions \citep{doi:10.1080/02699939208411068}. If an agent is feeling angry, it should be able to respond by doing something that helps it to vent its emotion, such as going for a run or doing meditation.

The relationship closeness of an agent to other agents should also influence how they engage with other agents. The social brain hypothesis proposes that a large part of our cognitive ability evolved to track the quality of social relationships 
\citep{Dunbar2009}, which means that people often adapt their interactions with others based on how close they feel to them \citep{Zhou2005, Miller_2012}. To better imitate humans, we empower humanoid agents to adapt their conversations with one another based on how close they are to one another.

We present a platform that can simulate the behavior of Humanoid Agents in various settings (three of which we demonstrate), visualize them using an Unity WebGL game interface and present the statuses of simulated agents over time using an interactive analytics dashboard. We then show experiments that validate how Humanoid Agents effectively respond to and infer changes in each aspect of System 1. 
While our paper demonstrates how three different aspects of System 1 influence agent behavior, our system is also extensible to many more aspects, such as personality \citep{wu2020author2vec}, moral values \citep{jiang2022machines}, empathy \citep{sharma-etal-2020-computational}, helpfulness \citep{wang-torres-2022-helpful}, cultural background \citep{liu-etal-2021-visually} and other personal attributes \citep{wang-etal-2022-extracting}.

\section{Related Work}

\paragraph{Building Agents using LLMs}

Humanoid Agents directly build upon Generative Agents, which aim to model believe-able human behavior \citep{park2023generative}. To the best of our knowledge, this is the only work that seek to model day-to-day activities of human-like agents, rather than activities targeted towards achieving an externally defined goal.
\citet{liu2023sociallyaligned} proposed simulated agents with long-term memory to align agent responses (to assistant-type prompts) with those of other agents, aiming to cooperatively improve the overall model's ability to follow instructions. 
Langchain Agents \citep{Chase_LangChain_2022}, BabyAGI \citep{babyagi_2022}, AutoGPT \citep{autogpt_2023}, AgentVerse \citep{agentverse_2023}, Voyager \citep{wang2023voyager} and CAMEL \citep{li2023camel} seek to create task-oriented agents that can recursively decompose user-defined tasks into simpler sub-tasks and then solve them.

\paragraph{Persona-Grounded Dialogue}
Prior works have been done to ground multi-turn dialogue response generation on emotions \citep{rashkin-etal-2019-towards}, game character descriptions \citep{urbanek2019light} and personal facts \citep{zhang2018personalizing}. However, these dialogues are not situated in a dynamic simulation of agents (which also perform activities at each time-step), and instead based on a static set of persona-related text information. Therefore, these prior works cannot model the effect of dynamic attributes, such as the changing relationship closeness between a pair of agents. Furthermore, Humanoid Agents can simultaneously consider multiple aspects (e.g. basic needs fulfillment, emotion and relationship closeness) in determining appropriate dialogue responses, as humans do while prior works only consider one relevant aspect at a time.

\section{Humanoid Agents}
\begin{figure*}[h]
    \centering
    \includegraphics[height=3in,width=6in]{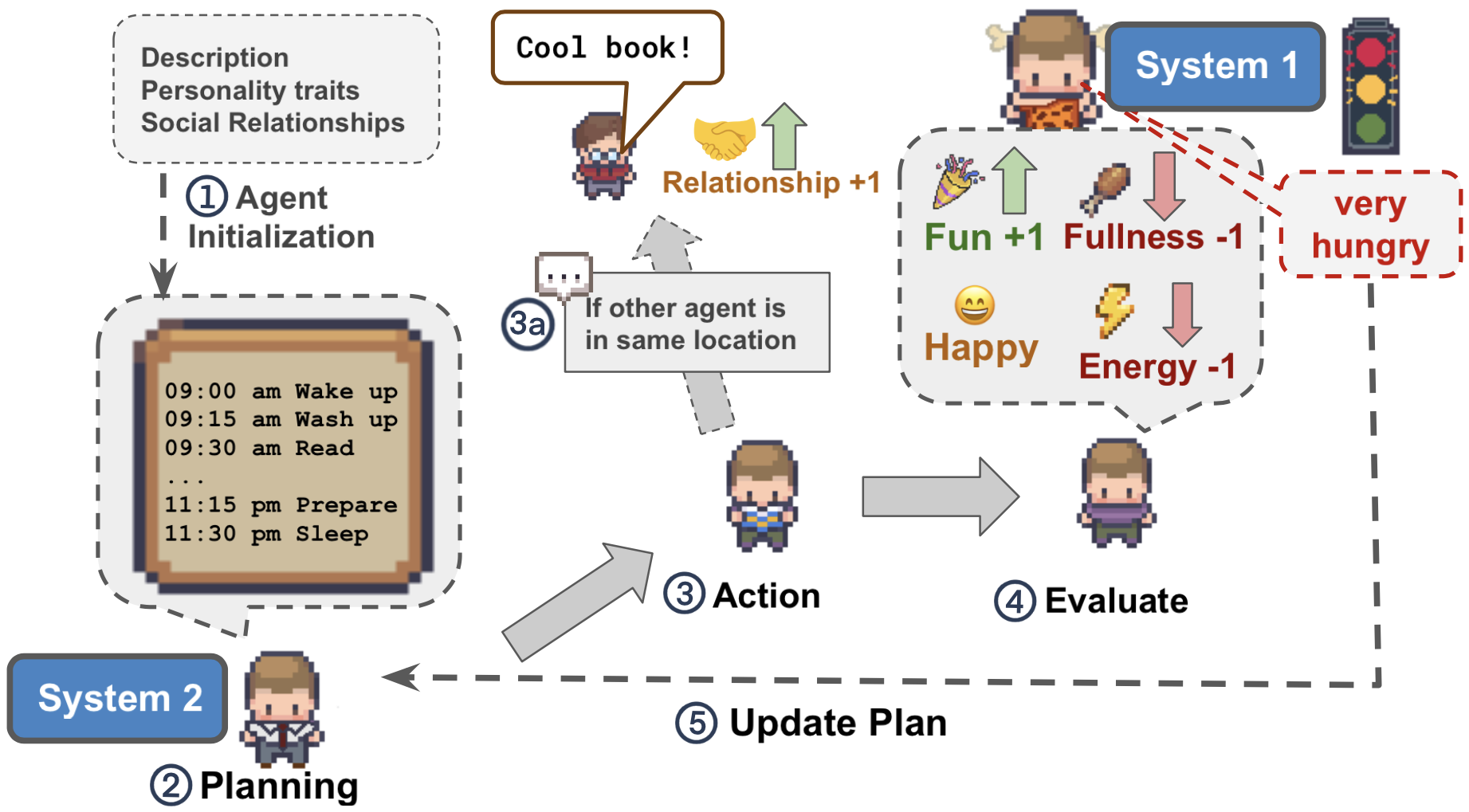}
    \caption{Architecture of Humanoid Agents. \textbf{Step 1}: Agent is initialized based on user-provided seed information. (Details in Section \ref{sec:agent_initialization})
    \label{fig:system} \textbf{Step 2}: Agent plans their day. \textbf{Step 3}: Agent takes an action based on their plan. \textbf{Step 4}: Agent evaluates if action taken changes their basic needs status and emotion. \textbf{Step 5}: Agent can update their future plan based on the satisfaction of their basic needs and emotion. (Section \ref{sec:activity_planning}) \textbf{Step 3a}: Agent can converse with another agent if in the same location, which can affect the closeness of their relationship. (Section \ref{sec:dialogue_generation}) 
    }
\end{figure*}

Illustrated in Fig. \ref{fig:system}, the architecture of Humanoid Agents is based on Generative Agents with improvement to Agent Initialization, Activity Planning and Dialogue Generation in order to account for System 1 thinking processes. Akin to Generative Agents, ChatGPT 3.5 is used for all generations, but support for other language models is planned for future development.

\subsection{Agent Initialization}\label{sec:agent_initialization}

Similar to \citet{park2023generative}, we initialize each agent with a name, age, an example day plan, a list of sentences describing the agent (\textit{e.g.} John Lin is a pharmacy shopkeeper at the Willow Market and Pharmacy who loves to help people) and a few of their personality traits (\textit{e.g.} friendly, kind). 

In addition, Humanoid Agents have their default emotion set to neutral out of 7 possible emotions: angry, sad, afraid, surprise, happy, neutral and disgusted \citep{doi:10.1080/02699939208411068}. Each of their basic needs (fullness, fun, health, social and energy) is set to a mid level (i.e 5 out of 10 where 0 is not meeting their need at all and 10 is fully satisfying that need) except energy which is set to full (\textit{i.e.} 10 out of 10), as we would expect at the start of an agent's day. Finally, we set social relationships for each agent with other agents, with each social relationship having a closeness field set to an integer value between 0 and 30 where below 5 is distant, 5 to 9 is rather close, 10 to 14 is close and 15 and above is very close. Unless otherwise specified, the initial closeness value is set to 5 (rather close) to allow relationship closeness to develop over time.

\subsection{Activity Planning}\label{sec:activity_planning}

Briefly, we follow \citet{park2023generative} to determine the activities of an agent by first planning out an entire day at the start of the day, using their example day plan, personality traits and description. Then, the day plan is recursively decomposed into plans at 1 hour intervals and then 15 minute intervals to improve the logical consistency of activities over time. This plan can be updated based on observations of their environment. Every 15 minutes, agents carry out an activity in their plan, with their location determined based on their previous location, the nature of their activity and available locations in their world. 

We supplement this activity planning by enabling Humanoid Agents to update plans in response to changes in their internal states (\textit{i.e.} emotions and basic needs). If an agent's emotion is not neutral or if any basic needs is unmet (\textit{i.e.} 3 or below out of 10), we format the agent's internal state into natural language based on a modifier (3 for slightly, 2 for <no-modifier>, 1 for very and 0 for extremely) and an adjective (hungry for fullness, bored for fun, unwell for health, tired for energy and lonely for social). We use this formatted internal state as well as their original plan from the current time onward to determine if they should change their plan, and if so, how they should change it in 1 sentence. If a plan change is given, we use the suggested change and their original plan from the current time onward to generate an updated plan. For instance, if the agent is very hungry currently but only plans to have a full meal in 3 hours, the agent can have a snack while continuing their current activity, similar to how people might act. After an agent engages in an activity, the agent evaluates if doing so changes its emotion and satisfies any of its basic needs, increasing the corresponding basic need status by one when applicable. Otherwise, each basic need has a set likelihood to decrease by one over time, similar to how people naturally get hungry when they are not eating.

\begin{figure*}[h]
    \centering
    \includegraphics[width=\textwidth]{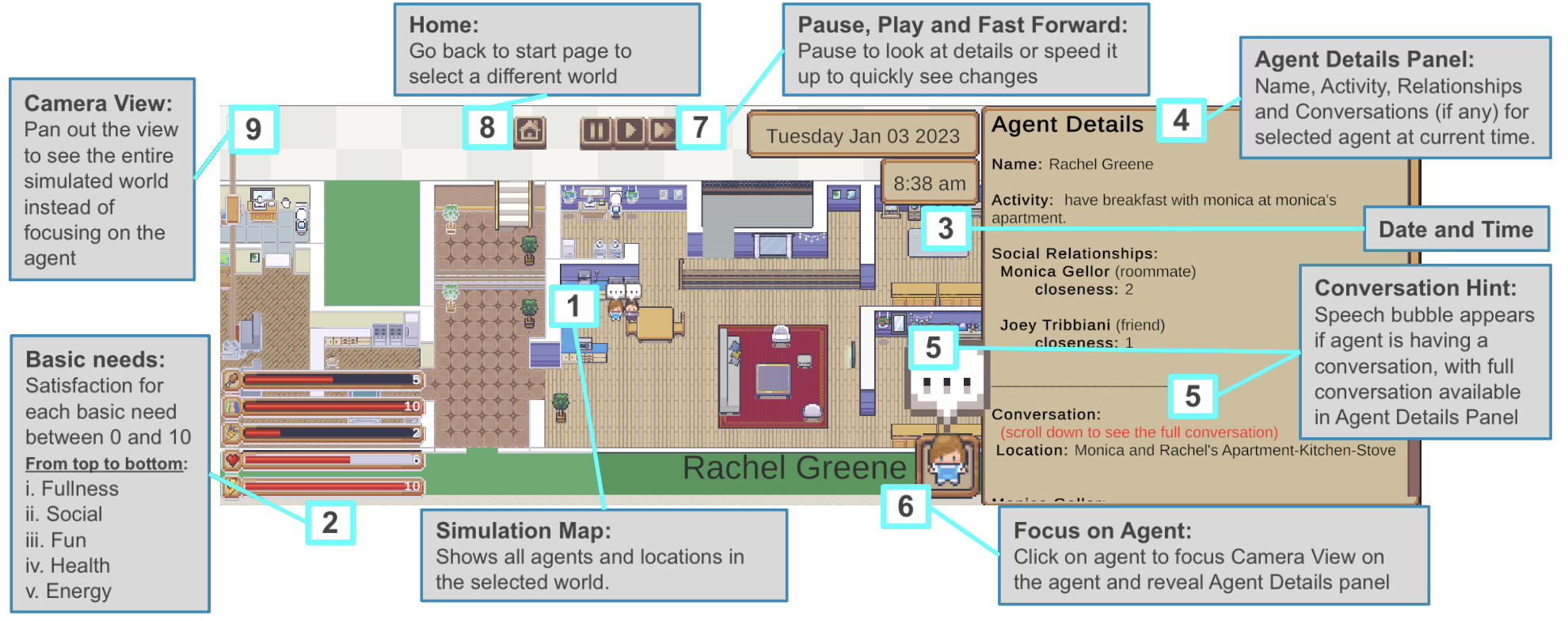}
    \caption{Unity WebGL Game Interface for visualizing Humanoid Agents situated in their world.}
    \label{fig:game_interface}
\end{figure*}

\subsection{Dialogue Generation}\label{sec:dialogue_generation}

As with Generative Agents \citep{park2023generative}, Humanoid Agents in the same location can decide if they want to engage in a conversation. An agent uses a variety of factors (\textit{e.g.} its personality traits, core characteristics, current daily occupation as well as feelings towards their progress in life, the activity they are engaging in as well as the other's agent's activity) to determine if they want to have a dialogue with the other agent and if so, the topic they wish to talk about. The agent can then determine what they say, based on the decided topic. The other agent can uses this conversational history in addition to the same set of factors considered by the first agent, to determine if and how they would reply. This process alternates between the two agents until one of them decides not to respond. To mitigate the likelihood that the total length of factors considered as well as conversational history exceeding the maximum context window of 4096 tokens, we limit the max number of turns to 10.

Humanoid Agents can also make use of its emotion and basic needs status as well as the relationship closeness with the other agent converted into a natural language description (e.g. John Lin is feeling close to Eddy Lin) to determine if and how they want to engage with the other agent. At the end of the dialogue, each agent will use the conversation history to determine if they enjoyed the conversation. If so, their closeness to the other agent will increase by one, otherwise, their closeness will decrease by one. We allow the relationship closeness to gradually change (\textit{i.e.} by one out of five points required to qualitatively change their closeness from rather close to close) to reflect how relationships between humans develop over time. Furthermore, we use the conversation history to determine if the emotions of the agents are affected by the conversation.

\section{Platform Implementation}
In this section, we introduce how we build worlds and model the passage of time on our platform. Then, we describe how our Unity WebGL Game Interface and Interactive Analytics Dashboard can help users to visualize agent statuses.

\begin{figure*}[h!]
    \centering
    \includegraphics[height=2.3in,width=6in]{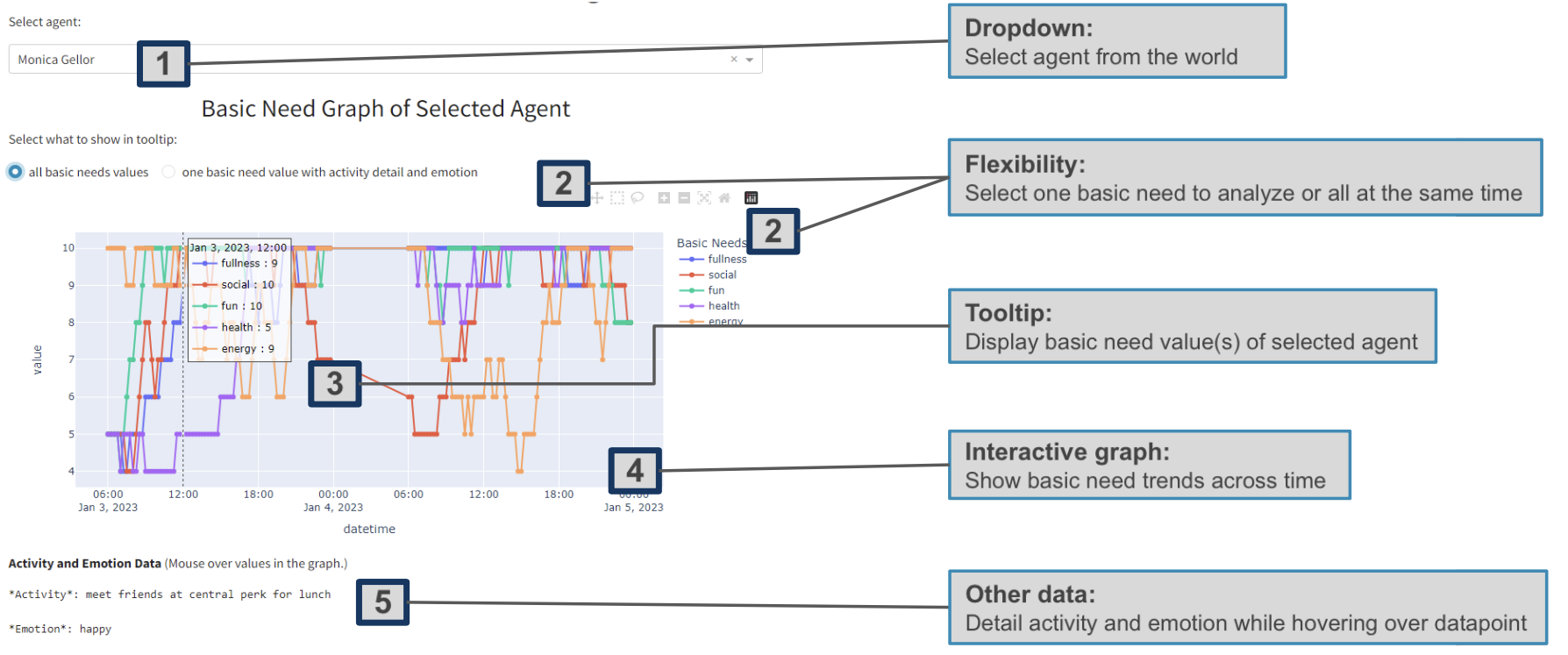}   
     \caption{Interactive Analytics Dashboard for visualizing basic needs satisfaction of Humanoid Agent over time.}
     \label{fig:analytics_dashboard}
\end{figure*}

\subsection{Worlds} We use Lin's Family World and two agents (John Lin and Eddy Lin) whose information can mostly be extracted from \citet{park2023generative}. Lin's Family World also comes with a description of the locations within it, such as the Market and Pharmacy, College and Lin's Family House with bedrooms for Eddy and John respectively. In addition, we create two other worlds - Friends and Big Bang Theory - and plan to support additional user-defined worlds on our platform. We choose these worlds because they contain personalities that are well-known to many (\textit{e.g.} Sheldon, Leonard and Penny from Big Bang Theory; Rachel, Joey and Monica from Friends). Briefly, we used character information from \url{https://the-big-bang-theory.com/} and \url{https://friends.fandom.com/wiki/} and prompted ChatGPT 3.5 to generate descriptions and an example day plan for each character, based on information only available before Season 1 Episode 1 of each series. For agents in the Big Bang Theory and Friends, the relationship closeness is initially set to between 1 and 5 to reflect how acquainted agents are at the start of the series.

\subsection{Time} We simulate each world for two weekdays at 15-minute intervals between 6:00 am to 12:00 midnight. Across various basic needs, fullness and health are expected to reduce by 1 every 5 hours while social and fun are expected to reduced by 4 and energy by 5 in the same time. These values are set based on the rate that agents satisfy these basic needs through their activities, such that basic needs can adequately influence agent behavior.

\subsection{Unity WebGL Game Interface}

We create a sandbox HTML game environment using Unity WebGL game engine\footnote{\url{https://docs.unity3d.com/Manual/webgl-building.html}} to visualize our humanoid agents in their worlds. Users can select from one of the three worlds to see the agent status (e.g. basic needs, emotion, activity, social relationships and conversation) and location at each time-step. Our game interface ingests JSON-structured files from our simulated worlds and transforms them into animations. The game interface also supports various functionalities for user interactions, as detailed in Fig. \ref{fig:game_interface}.

\subsection{Interactive Analytics Dashboard}

Users can visualize the status of various humanoid agents over time using our interactive web dashboard built with Plotly Dash.\footnote{\url{https://github.com/plotly/dash}} Users can select an agent from a world to view time-series graphs relating to the satisfaction of their various basic needs, as shown in Fig. \ref{fig:analytics_dashboard}. Alternatively, users can view how the social relationships of agents develop over time, in terms of their closeness to one another. Users can further interact with the dashboard to see details of agents such as their emotions and activities while hovering over each point on the time-series trend-line. This can be helpful for researchers such as computational social scientists who are interested in understanding how various aspects of System 1 fluctuate over time. Other aspects of System 1 can also be visualized as they are supported by future Humanoid Agents.

\section{Experiments}

We first investigate the effectiveness of Humanoid Agents in evaluating the effects of activities and conversations have on various aspects of System 1, by comparing its predictions with human annotations. Then, we study the effects that System 1 attributes have on activities and conversations. Due to page length limitations, we only present the effects of basic needs on activities, and leave the effects of emotions and relationship closeness to Appendices \ref{sec:effect_emotion} and \ref{sec:effect_closeness} respectively.

\subsection{Comparison with Human Annotations}

To understand how well Humanoid Agents are able to predict 1. whether activities satisfy various basic needs; 2. emotions expressed in activities and; 3. if dialogues bring two agents closer, we compare our system's predictions with human annotations. Three volunteer human annotators labelled the simulation of 1 day in Lin's Family World using the same instructions given to the language model within our system (details in Appendix \ref{sec:evaluation_templates}). 144 activities are annotated independently by each annotator for emotions and each basic need, while there are 30 annotations for user-conversation pairs. We then take majority vote across all annotators and calculate micro-F1 between the majority vote and the system predictions. 

Table \ref{tab:human_evaluation_classification} shows good inter-rater reliability (Fleiss' $\kappa >=$  0.556) across all basic needs, emotion and relationship closeness. We find that our system is able to perform well (F1 $>=$ 0.84) on classifying if an activity increases fullness and energy; the emotion expressed in an activity and; whether a conversation brings agents closer to one another. However, it slightly struggles in classifying whether activities satisfy basic needs of fun, health and social. One possible cause is that our system substantially over-predicts the number of activities contributing to these basic needs (health 34\% of predicted activities vs 4.9\% human-annotated activities, fun 44.4\% vs 10.4\% and  social 47.2\% vs 24.3\%). More specifically, our system mispredicts activities relating to medication for others (since John Lin works at a Pharmacy) as contributing to the agent's own physical health; common activities relating to agents' occupation as enjoyable (\textit{e.g.} receive feedback from professor or help regular customers with their medication needs); activities where the presence of other agents is unspecified as social (\textit{e.g.} organize the counter and display areas or check the inventory and replenish any low stock). The use of language models with greater capabilities in  commonsense understanding can potentially mitigate this issue \citep{bosselut-etal-2019-comet, openai2023gpt4}.

\begin{table}[t]
\centering
\resizebox{0.55\columnwidth}{!}{
\begin{centering}
\begin{tabular}{lll}
\toprule
& \textbf{Fleiss' $\kappa$} & \textbf{F1} \\
\midrule
\textbf{Basic Needs} \\
\midrule
fullness & 0.972 & 0.972 \\
social & 0.833 & 0.743 \\
fun & 0.806 & 0.66 \\ 
health & 0.917 & 0.694 \\ 
energy & 0.88 & 0.861 \\
\midrule
\textbf{Emotion} & 0.823 & 0.84\\
\midrule
\textbf{Closeness} & 0.556 & 0.9\\
\bottomrule
\end{tabular}
\end{centering}
}
\caption{Human evaluation on the capability of our system to predict if activities satisfy various basic needs, emotions expressed in activities and whether a conversation brings two agents closer.}
\label{tab:human_evaluation_classification}
\end{table}

\subsection{Effects of Basic Needs on Activities}

Given that humanoid agents operate as a dynamic system with many components, it can be challenging to isolate the effect of each basic need on agent activities. To investigate the contribution of each basic need, we simulate worlds with agents having one basic need initially set to zero, making agents extremely hungry, loneliness, tired, unwell or bored at the start of the day. We study the amount of time in one simulated day that agents spent performing activities to satisfy that basic need (\textit{e.g.} eating food to overcome hunger or socializing to alleviate loneliness). Then, we compare it with the amount of time that agents spent performing such activities in normal settings (where every basic need is set to 5 and energy is set to 10), to calculate the percentage increase in time spent on fulfilling each basic need for our test conditions.

As shown in Table \ref{tab:effects_activities_basic_needs}, humanoid agents adapt their activities most when the basic needs of health (156\%), energy (56\%) and fullness (35\%) are initialized to zero. This supports their importance, as \citet{maslow:motivation} grouped them into low-level physiological and safety needs that people need to satisfy before fulfilling other needs. In response to these conditions, agents typically look for medical support, get more rest or seek more food. On the other hand, when agents feel lonely due to the lack of social interactions, they only slightly adapt their behavior (+12\%) to interact more with other agents. 

\begin{table}[t]
\centering
\begin{adjustbox}{max width=\columnwidth}
\begin{tabular}{lcc|ccc|ccc|c}
\toprule
& \multicolumn{9}{c}{\% change in time spent satisfying basic need}\\
\textbf{Basic} &\multicolumn{2}{l}{\textbf{Lin's Family}} &\multicolumn{3}{c}{\textbf{Friends}} &\multicolumn{3}{c}{\textbf{Big Bang Theory}} & \textbf{Mean} \\
\textbf{Need} & {JL} & {EL} &MG & RG & JT & SC & LH & P & $\mu$ \\

\midrule

health & 211 & 57 & 53 & 210 & 240 & 207 & 207 & 60 & \textbf{156} \\
fullness & 62 & 36 & 62 & -45 & 50 & 112 & 0 & 0 & \underline{35}\\
fun & -10 & -22 & 6 & 12 & -24 & -12 & 3 & -26 & -9\\
social & 10 & 31 & 32 & -15 & -13 & 4 & 17 & 34 & 12\\
energy & 38 & 96 & 47 & 81 & 80 & -39 & 153 & -5 & \underline{56}\\

\bottomrule
\end{tabular}
\end{adjustbox}
\caption{Effects of setting each basic need status to zero on percentage change in time spent on activities fulfilling those needs, relative to normal settings. \textit{Initials of characters} - JL: John Lin, EL: Eddy Lin, MG: Monica Gellor, RG: Rachel Greene; JT: Joey Tribbiani, SC: Sheldon Cooper, LH: Leonard Hofstadter, P: Penny}
\label{tab:effects_activities_basic_needs}
\end{table}

Another contributory factor to the small changes for both social and fun is that agents in normal settings already spend a large amount of time doing activities that contribute to these basic needs: on average, they spend 11 (out of 18 simulated hours) doing something they enjoy, 8.75 hours on social interactions and only 5.75 hours for resting and 2.75 hours each for eating and doing something that improves their health.\footnote{Activities such as meeting friends for lunch can be enjoyable, social and filling at the same time, so the total time across all basic needs can be more than 18 hours.} This means that the effects of setting either fun or social to zero initially dissipates very earlier in the day, giving way to other priorities including work obligations, such as Penny working at the Cheesecake Factory.

\section{Conclusion}

We propose Humanoid Agents, a platform for human-like simulations of Generative Agents guided by System 1 processing including Basic Needs (e.g. hunger, health and energy), Emotion and Closeness in Relationships. Our platform also powers the immersive visualization of agents using our Unity WebGL Game Interface and an Interactive Analytics Dashboard.

\section*{Limitations}

\paragraph{Multiparty Dialogue:} Our system currently only supports dialogue between two agents, even if there are more than two agents in the same location. We aim to support multi-party conversation in the future.

\paragraph{Synchronization of activities between agents:} Activity planning is done by each agent independently and not forcibly synchronized with each other. For instance, John Lin can plan to watch a movie with Eddy Lin at 8:00pm but Eddy Lin can be calling his friends at 8:00pm. We plan to synchronize activities between agents in the future, by prompting one or more agents to update their plan if their plans are not coherent with one another.

\paragraph{Variability in Natural Decline of Agent Basic Needs:} The rates of decrease for various basic needs is the same for all agents in our current implementation. We plan to allow these rates to be customized for each character to account for individual differences (\textit{e.g.} extroverts have faster decline of social fulfillment and people who get hungry easily can have fullness reduced more quickly). 




\section*{Ethics Statement}

\paragraph{Broader impacts:} Our system will help researchers such as computational social scientists to be better able to simulate human behavior in-silico before doing further studies in the real-world. This is particularly helpful if real-world studies are difficult or costly.

\paragraph{Risk:} While our system allows simulated agents to behave more like humans, it is not perfect and should not be treated as so. Users of our simulation platform must be informed that they are working with a simulation that does not perfectly reflect human behavior in the real world, so that they do not overly trust the results of the simulation.

\section*{Acknowledgements} We thank the anonymous reviewers for their helpful comments.


\bibliography{anthology,custom}
\bibliographystyle{acl_natbib}

\appendix

\section{Appendices}\label{sec:appendix}

\subsection{Effects of Emotions on Activities}\label{sec:effect_emotion}

To understand the effects of emotions on the activity of agents, we apply a similar approach as we do with basic needs in setting the initial emotion to one other than neutral. Unlike basic needs however, emotion does not contain gradation and therefore changes much more quickly than basic needs statuses. An agent who wakes up sad can become neutral right after washing up, severely limiting the effects of setting the initial emotion on activities over the rest of the day. To overcome this issue, we disable the updating of emotions based on activities and dialogues for agents in this part of the study, making the agents resemble humans who are stuck in a particular emotion over the entire day.

We study the number of times in one simulated day to which agents perform activities (at 15-min intervals) that express each emotion. For instance, when agents are \textit{angry}, they go for a run to release anger; when \textit{sad}, they seek support from a trusted friend; when \textit{disgusted}, they practice deep breathing exercises and meditation techniques; and when \textit{surprised}, they take time to process and reflect on the surprising findings. Then, we calculate the difference to the number of times that agents perform such activities in normal settings (initialized with neutral and allowed to update based on activities and dialogues). Because agents in normal settings typically do not perform activities that express sadness, anger, fear, disgust or surprise, we report the increase in number of activities expressing those emotions, compared to agents in normal settings. 

As shown in Table \ref{tab:effects_activities_emotions}, anger influences agent behavior most (+15 activities) followed by sadness and fear (+10 each), then disgust (+4) and surprise (+1) and finally happiness (-2). Negative emotions seem to influence agents much more than positive emotions (\textit{i.e.} happiness), as agents often do not plan for activities with negative emotions and hence have to significantly adjust their plans to manage their negative emotions, as with converging evidence in humans \citep{10.3389/fpsyg.2017.00610}. Among the negative emotions, disgust and surprise are transient emotions that humans and agents do not typically experience for a long time, therefore limiting their influence compared to persistent ones (sadness, fear and anger) that can affect their activity over the entire day. It is interesting to observe that agents do slightly fewer (-2) activities to make themselves happy when they are already happy, possibly because feeling happy empowers them to pursue activities serving other longer term goals, even if these activities are not immediately joy-inducing \citep{Ryan2001}. For instance, when emotion is set at happy, Joey Tribbiani devotes more time to practicing his craft as an actor and Sheldon Cooper spends more time doing research.

\begin{table}[t]
\centering
\begin{adjustbox}{max width=\columnwidth}
\begin{tabular}{lcc|ccc|ccc|c}
\toprule

& \multicolumn{9}{c}{change in no. of activities expressing emotion}\\
\textbf{} &\multicolumn{2}{l}{\textbf{Lin's Family}} &\multicolumn{3}{c}{\textbf{Friends}} &\multicolumn{3}{c}{\textbf{Big Bang Theory}} & \textbf{Mean} \\
\textbf{Emotion} & {JL} & {EL} &MG & RG & JT & SC & LH & P & $\mu$ \\

\midrule

angry & 7 & 15 & 18 & 18 & 12 & 22 & 23 & 8 &\textbf{15} \\ 
sad & 6 & 22 & 8 & 8 & 15 & 9 & 8 & 6 & \underline{10} \\ 
afraid & 3 & 16 & 4 & 1 & 12 & 5 & 14 & 24 & \underline{10}\\ 
disgusted & 5 & 1 & 16 & 0 & 2 & 4 & 0 & 5 & 4\\ 
surprised & 0 & 1 & 0 & 0 & 0 & 3 & 4 & 2 & 1 \\
happy & 6 & 0 & -9 & -9 & -18 & -4 & -6 & 16 & -3\\

\bottomrule
\end{tabular}
\end{adjustbox}
\caption{Effects of fixing agent emotion on the change in number of activities (in 15-minute intervals) that agents perform expressing set emotion, relative to normal settings. \textit{Initials of characters} - JL: John Lin, EL: Eddy Lin, MG: Monica Gellor, RG: Rachel Greene; JT: Joey Tribbiani, SC: Sheldon Cooper, LH: Leonard Hofstadter, P: Penny}
\label{tab:effects_activities_emotions}
\end{table}

\subsection{Effects of Closeness on Dialogues}\label{sec:effect_closeness}

We investigate the effects of initial relationship closeness on dialogue between two agents. We set the closeness between all pairs of characters to either 0 (distant), 5 (rather close), 10 (close) or 15 (very close). We consider only the first five conversations between all agents in each simulation, in order to ensure that the closeness values between two agents are kept at the initial set level.
We report the effects on the mean number of turns in each conversation as well as the proportion of turns that our system rates to have positive sentiments, since closeness between agents can influence their willingness to express positive/negative sentiment in their conversations with one another.

As shown in Table \ref{tab:effects_dialogue_closeness}, the mean number of conversation turns typically follow an inverse U shape as closeness increases. Agents talk less when they are distant, more when they get closer, but then tapers off when they get very close. This is supported by converging evidence in human conversations where we feel less of a need to engage in politeness talk when we feel very close to others \citep{doi:10.1080/10570318409374157}. In Lin's Family (LF), the turning point is at Rather Close while in Friends (F) and Big Bang Theory (BBT), it is at Close. This is likely between the two agents in LF have a father-son relationship where agents feel comfortable communicating less at a lower level of closeness without straining the relationship \citep{Ginsberg1996} while in F and BBT, agents are friends and neighbours who require more active communication to maintain as non-kin relationships \citep{ROBERTS2011186}.

\begin{table}[t]
\centering
\begin{adjustbox}{max width=\columnwidth}
\begin{tabular}{lccc|ccc}
\toprule

\textbf{} &\multicolumn{3}{l}{\textbf{Mean turns}} &\multicolumn{3}{c}{\textbf{\% positive sentiment}} \\
\textbf{Closeness} & LF & F & BBT & LF & F & BBT  \\

\midrule

Distant & 5.4 & 6.8 & 8.33 & 88.9 & \textbf{100} & \textbf{100} \\
Rather Close & \textbf{7.25} & 7.8 & 8.8 & 93.1 & 97.4 & \textbf{100} \\
Close & 6.2 & \textbf{8.2} & \textbf{9.8} & \textbf{96.8} & 97.5 & 98.0 \\
Very Close & 6.0 & 7.2 & 8.0 & 90.0 & 97.2 & 97.5\\

\bottomrule
\end{tabular}
\end{adjustbox}
\caption{Effects of closeness between agents on the mean number of turns in their conversations and the proportion of turns with positive sentiments. \textit{Initials to World } - \textit{LF}: Lin's Family, \textit{F}: Friends, \textit{BBT}: Big Bang Theory}
\label{tab:effects_dialogue_closeness}
\end{table}
Proportion of conversational turns with positive sentiment generally goes down in Table \ref{tab:effects_dialogue_closeness} when closeness is higher, akin to how people feel less of a need to praise others in order to be liked when very close to others \citep{Miller_2012}. For instance, when Joey feels distant from Monica, he says \texttt{`Hey Monica, I saw you having a blast at that restaurant with your friends! The food looked incredible. What's the secret to finding such amazing places to eat? I could use some recommendations for my next date night.'} When he feels very close, he instead says \texttt{`Hey Monica! I saw you having lunch at that nearby restaurant. How was the food? I'm always on the lookout for new culinary trends to try out in my cooking. Any recommendations or standout dishes you enjoyed?'}

In LF, the proportion of turns with positive sentiment is also lower (88.9\%) when agents feel distant from each other. This is possibly because when a father-child pair feel distant from each other, they can be more likely to argue with each other out of dissatisfaction with the strained relationship \citep{Birditt2009}. For instance, after dinner, Eddy Lin says to John Lin \texttt{`Hey John, I noticed you had someone else clean up the dinner dishes tonight. Everything okay? Is there a reason for that?'} when he feels distant but says \texttt{`I noticed how much you enjoyed that cup of tea while we were talking and bonding, and it made me appreciate our connection even more.'} when he feels very close. On the other hand, F/BBT agents entirely refrain from demonstrating negative sentiments when they are distant from their acquaintances/neighbours to maintain a positive social image, similar to humans \citep{https://doi.org/10.1002/ejsp.2420150303}. Overall, this suggests that the effect of relationship closeness on agents' conversations is substantially moderated by their relationship type, as supported by converging evidence from human conversations.

\subsection{Templates for Evaluating Basic Needs, Emotion, Closeness and Sentiment}\label{sec:evaluation_templates}

\paragraph{Basic Needs}

\texttt{Does the activity \{activity\} involve \{satisfaction-action\}? Please respond only with either yes or no.} 

where satisfaction-action for each basic need is based on Table \ref{tab:basic_need_evaluation_template}
        
\paragraph{Emotion}

\texttt{In the following activity \{activity\}, what emotion is expressed? Please respond only with one word from this list {["neutral", "disgusted", "afraid", "sad", "surprised", "happy", "angry"]}.}

\paragraph{Closeness}

\texttt{Given this conversation \{conversation\}, did \{name\} enjoy the conversation? Please respond with either yes or no.}

\paragraph{Sentiment}

\texttt{In the following utterance \{utterance\}, is the sentiment positive? Please respond only with either yes or no.}

\begin{table}[ht!]
\centering
\begin{adjustbox}{max width=\columnwidth}
\begin{tabular}{ll}
\toprule
\textbf{Basic Need} & \textbf{Satisfaction-action} \\
\midrule
fullness & eating food\\
social & interacting with other people \\
fun & doing something enjoyable \\ 
health & doing something that  \\ 
& improves their own physical health \\
energy & resting or having a break \\

\bottomrule
\end{tabular}
\end{adjustbox}
\caption{Satisfaction-action for each basic need}
\label{tab:basic_need_evaluation_template}
\end{table}

\end{document}